\documentclass{article}

\usepackage{microtype}
\usepackage{graphicx}
\usepackage{subcaption}
\usepackage{booktabs} 
\usepackage{float}    
\usepackage{hyperref}

\usepackage[preprint]{icml2026}
\hypersetup{pdfsubject={}, pdftitle={HiSA: Hidden Sparse Attention},
  pdfauthor={Ambroise Borbely, Mathieu Salzmann, Olivier Verscheure}}

\usepackage{amsmath}
\usepackage{amssymb}
\usepackage{mathtools}
\usepackage{amsthm}
\usepackage[capitalize,noabbrev]{cleveref}

\theoremstyle{plain}

\theoremstyle{definition}

\usepackage[textsize=tiny]{todonotes}

\newcommand{\method}{HiSA}
\newcommand{\pairq}[1]{\mathbf{q}^{(#1)}}
\newcommand{\pairk}[1]{\mathbf{k}^{(#1)}}
\newcommand{\Phimax}{\Phi}
\DeclareMathOperator{\softmax}{softmax}

\icmltitlerunning{HiSA: Sparsifying the Hidden Dimension of RoPE Attention in LLMs}

\begin{document}

\twocolumn[
  \icmltitle{HiSA: Sparsifying the Hidden Dimension of RoPE Attention in LLMs}

  \icmlsetsymbol{corr}{*}

  \begin{icmlauthorlist}
    \icmlauthor{Ambroise Borbely}{corr,epfl,sdsc}
    \icmlauthor{Mathieu Salzmann}{epfl}
    \icmlauthor{Olivier Verscheure}{sdsc}
  \end{icmlauthorlist}

  \icmlaffiliation{epfl}{EPFL, Switzerland}
  \icmlaffiliation{sdsc}{Swiss Data Science Center, EPFL and ETH Zurich, Switzerland}

  \icmlcorrespondingauthor{Ambroise Borbely}{ambroise.borbely@epfl.ch}

  \icmlkeywords{Large Language Models, Sparse Attention, Rotary Position Embedding, Efficient Attention, KV Cache,
    Long Context}

  \vskip 0.3in
]

\printAffiliationsAndNotice{\textsuperscript{*}Corresponding author. }

\begin{abstract}
  Attention in large language models (LLMs) scores every query against every key with a full dot product. We investigate whether this dot product must be computed completely for attention to be effective. We propose \emph{Hidden Sparse Attention} (\method{}), which keeps the query and key projections intact but only computes a partial dot product, following a static, parameter-free rule set by the token positions: attention becomes sparse in the hidden (feature) dimension, an axis much less explored than token sparsity, rather than over tokens as in almost all sparse attention methods. Based on the structure of rotary position embeddings (RoPE), we decompose keys and queries into parts which can be omitted when computing attention. By training LLMs from scratch with \method{}, we find that this partial dot product does not hurt but improves performance compared with the complete product while computing only about a third of the query--key products. During inference \method{} improves decoding speed and shrinks memory footprint. We also extend \method{} to FlashAttention-3 kernels for training and inference. Finally, we motivate our approach by showing that dense attention learns to suppress the attention score parts that \method{} drops. Code is available at \url{https://github.com/ambroise-b/hisa}.
\end{abstract}

\section{Introduction}
\label{sec:intro}

The attention layer \citep{vaswani2017attention} of a large language model compares every query with every key via a dot product over the full head dimension. Most work on making attention cheaper keeps this dot product intact and instead reduces the set of keys a query looks at. Strategies to do so include fixed patterns  \citep{child2019sparse}, sliding windows \citep{jiang2023mistral}, KV-cache eviction \citep{xiao2023streamingllm,zhang2023h2o}, and query-dependent token selection, training-free \citep{tang2024quest} or learned \citep{yuan2025nsa}. In this work, we investigate an orthogonal question: whether attention requires computing the complete query--key dot product.

Rotary position embeddings (RoPE) \citep{su2021roformer}, used by many modern LLMs ~\citep{grattafiori2024llama3,yang2025qwen3,olmo2025olmo3, deepseekai2026deepseekv4,gemmateam2026gemma4}, give this question a natural structure. RoPE rotates each two-dimensional frequency pair of the query and the key by an angle proportional to its position. Because rotations compose, the two rotations combine into a single one that encodes the relative distance between the query and the key. High-frequency pairs complete many turns within a few tokens, whereas low-frequency pairs barely rotate over the whole context. 

The score parts of fast pairs are therefore highly sensitive to the exact distance between the query and the key, whereas language does not follow an exact structure over distant words or tokens, and sequences can be long. This raises the question of whether every pair needs to be computed at every relative distance. We show that high-frequency pairs are not necessary for the model when distance between tokens is large, and that removing them is harmless or even beneficial.

Our Hidden Sparse Attention (\method{}) method turns this hypothesis into an attention rule: the dot product of each frequency pair $l$ between a query and a key is kept only when the rotation induced by RoPE is below an angle budget $\Phimax$, a hyperparameter that we set to 90$^\circ$ by default. The query and key projections are unchanged and what becomes sparse is the hidden dimension of the dot product. The sparsity pattern follows from the RoPE frequencies and the positions alone: it has no learned parameter and needs no selection step, unlike sparse attention methods that score, route or select keys at runtime. For a given query, fast pairs only see nearby keys, slow pairs see far.

\begin{figure*}[t]
  \centering
  \includegraphics[width=\textwidth]{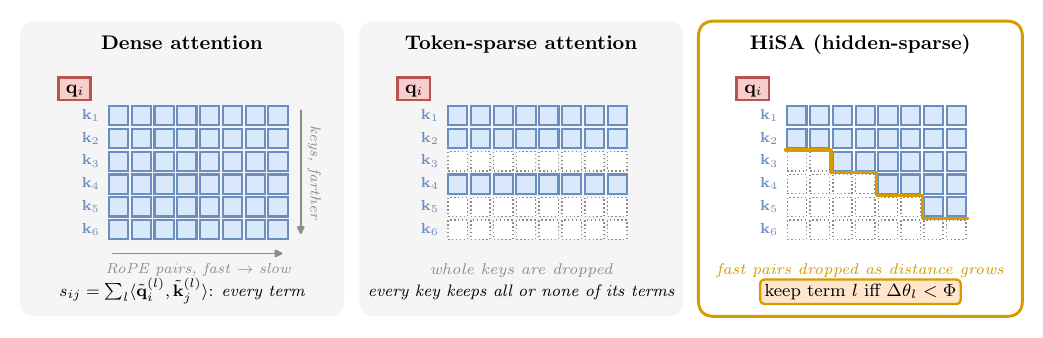}
  \caption{Dense attention computes every part
    (left); token-sparse attention drops whole keys (centre). \method{} keeps a part only while its rotation induced by RoPE is below an angle budget $\Phimax$: nearby keys are compared on all pairs, farther keys only on the slow ones, so attention becomes sparse in the hidden dimension rather than over tokens (right).}
  \label{fig:teaser}
\end{figure*}

We propose \method{}, a new sparse attention architecture that is sparse in the hidden dimension rather than over tokens. Our contributions can be summarized as follows:
\par\vspace{-4pt}

\begin{itemize}
  \setlength{\itemsep}{1pt}\setlength{\parsep}{0pt}\setlength{\parskip}{0pt}
  \item Based on the structure of RoPE, we develop \method{}, \textbf{a sparse attention in the hidden dimension} that computes a partial query--key dot product. Its sparsity pattern is \textbf{static and parameter-free}. We derive its compute, reach, and KV-cache footprint (\cref{sec:method}).
  \item We \textbf{evaluate \method{} in pretraining from scratch and in long-context extension}, implement it in FlashAttention for training and inference, and compare it with dense attention on validation loss, training speed, KV-cache memory, and decoding throughput (\cref{sec:exp-4k,sec:exp-fa,sec:exp-long,sec:exp-kv}).
  \item For existing models, we show that \textbf{a pretrained dense model can be converted} to \method{} by continued training and matches dense continued training (\cref{sec:exp-convert}).
  \item From an interpretability perspective, we show that \textbf{the score parts that \method{} drops correspond to those that dense attention learns to suppress} during training (\cref{sec:interp}): the rule is not only motivated by the geometry of RoPE, but measured in what dense models learn.
\end{itemize}
\section{Related Work}
\label{sec:related}

\paragraph{Attention and RoPE.}
Attention \citep{vaswani2017attention} with RoPE \citep{su2021roformer} encodes relative positions by rotating query and key pairs at spaced frequencies $\theta_l = b^{-2l/d}$. \citet{barbero2024round} show that Gemma 7B uses the highest frequencies for positional attention patterns and suggest that the lowest carry semantic information; \citet{men2024base} and \citet{liu2023scaling} relate the RoPE base to the usable context length, and \citet{wang2026adarope} learn per-head frequencies. Exact attention is made fast by the IO-aware FlashAttention kernels \citep{dao2022flashattention,dao2023flashattention2,shah2024flashattention3}, which we extend.

\paragraph{Sparse attention over tokens.} Static patterns and sliding windows \citep{child2019sparse,beltagy2020longformer,zaheer2020bigbird,jiang2023mistral} restrict each query to a subset of keys. For inference, KV-cache eviction keeps attention sinks and recent tokens \citep{xiao2023streamingllm} or heavy hitters \citep{zhang2023h2o}, and Quest \citep{tang2024quest} selects cache pages per query. Recent methods are trained with the sparsity in place, using block-sparse kernels: NSA \citep{yuan2025nsa} combines compressed, selected and sliding-window branches, and MoBA \citep{lu2025moba} routes each query to the top-scoring blocks of keys.

\paragraph{Sparsity in the hidden dimension.}
Several inference methods exploit the feature dimension, but mostly as a proxy for selecting tokens. SparQ \citep{ribar2023sparq} and Loki \citep{singhania2024loki} approximate the scores with a few query components or a low-rank key projection, select the top keys, and then compute the full dot product over them; Double Sparsity \citep{yang2024doublesparsity} calibrates important channels offline and FASA \citep{wang2026fasa} uses a few dominant RoPE frequency chunks for the same purpose. The final attention of these methods is sparse over tokens, not in the hidden dimension. ThinK \citep{xu2024think} does compute the final score on fewer features, by pruning key-cache channels, but chooses them from the content and applies the same pruning to all but the most recent tokens. All of these methods are applied after training. Sparse Feature Attention \citep{xie2026sfa} trains models whose queries and keys keep only their $k$ largest-magnitude features, chosen per vector from the content; \method{} instead fixes the kept parts by the positions alone, through the RoPE rotation angles.

Two concurrent works also restrict each RoPE pair to a window proportional to its wavelength. ATFlash \citep{hayashi2026atflash} applies such windows to pretrained models without training (two periods by default) and ports them to FlashAttention-4 prefill and FlashInfer decode, with speedups that grow with context length. WavePrune \citep{du2026waveprune} restricts each pair to one rotation period, both training-free and when pretraining GPT-2 models from scratch, with FlashAttention-based forward and backward kernels. WavePrune explains the window by the aliasing that fast pairs cause in the attention maps of pretrained models. Both works skip computation and key reads, but neither reduces the KV cache: every key and value stays in memory. \method{} differs in three ways. It trains with much tighter budgets (a quarter period at 90$^\circ$), where the partial product clearly improves the loss at the training length. It turns the rule into an actual KV-cache reduction, for keys per chunk and for values beyond the reach. Moreover, an interpretability analysis of dense models explains why the dropped pairs are not needed: dense training itself learns to suppress them.

\paragraph{Long-context extension.}
RoPE models are extended by interpolating positions \citep{chen2023pi,peng2023yarn} or raising the base
\citep{xiong2023effective}, followed by continued pretraining on long documents mixed with short data
\citep{xiong2023effective,gao2024prolong}. We follow this recipe to study \method{} beyond its pretraining length.

\section{Hidden Sparse Attention}
\label{sec:method}

\subsection{The RoPE score as a sum over frequency pairs}

Consider one attention head with dimension $d$ and $P = d/2$ RoPE pairs. Let $\pairq{l}_i, \pairk{l}_j \in \mathbb{R}^2$ define a pair $l$ consisting of a query at position $i$ and a key at position $j$ \emph{before} rotation, and $\theta_l = b^{-2l/d}$ be its frequency, with RoPE base $b$. 
RoPE rotates $\pairq{l}_i$ and $\pairk{l}_j$ by the angle $i\theta_l$ and $j\theta_l$, respectively.
Since rotations preserve norms and only the difference of the two rotations remains in the inner product, the attention score $s_{ij}$ decomposes exactly as
\begin{equation}
  s_{ij} = \frac{1}{\sqrt{d}} \sum_{l=0}^{P-1} \big\|\pairq{l}_i\big\| \, \big\|\pairk{l}_j\big\|
  \cos\!\big(\varphi^{(l)}_{ij} + \Delta\,\theta_l\big),
  \label{eq:decomp}
\end{equation}
where $\Delta = i - j$ and $\varphi^{(l)}_{ij}$ is the angle between $\pairq{l}_i$ and $\pairk{l}_j$. The \emph{content} of the query and the key enters through the norms and through $\varphi^{(l)}_{ij}$, whereas their \emph{position} enters only through the rotation $\Delta\theta_l$. We can also define each attention score part $s^{(l)}_{ij}$ as
\begin{equation}
  s^{(l)}_{ij} = \frac{1}{\sqrt{d}} \big\|\pairq{l}_i\big\| \, \big\|\pairk{l}_j\big\|
  \cos\!\big(\varphi^{(l)}_{ij} + \Delta\,\theta_l\big),
  \label{eq:part}
\end{equation}
so that $s_{ij} = \sum_{l} s^{(l)}_{ij}$.

Consider a fixed query and a key far from it. For a fast pair, moving the key by a single token, a negligible change relative to the distance $\Delta$, already rotates the pair by $\theta_l$ (up to one radian), so its score part $s^{(l)}_{ij}$ (\cref{eq:part}) changes drastically. Such a pair can only express \emph{exact} long-range positions. In other words, even a minor change in sentence structure or word order in the distant context of the key may have an important impact on the score. Language models, however, use distant context differently from nearby context: recurrent language models rely on word order only within the last few dozen tokens and treat more distant history as a coarse, order-insensitive context \citep{khandelwal2018sharp}. The relevant context far from the query is thus rarely tied to an exact offset. 

We therefore hypothesise that models do not need precise positional patterns at long range, which is exactly what fast pairs encode there. Slow pairs, on the other hand, rotate little over long distances and can still capture long-range dependencies, so they should be kept up to much larger distances. To test this hypothesis we define the \method{} rule: a new attention formulation which omits the score parts $s^{(l)}_{ij}$ when $\Delta\theta_l$ is large.

\begin{figure}[t]
  \centering
  \includegraphics[width=\columnwidth]{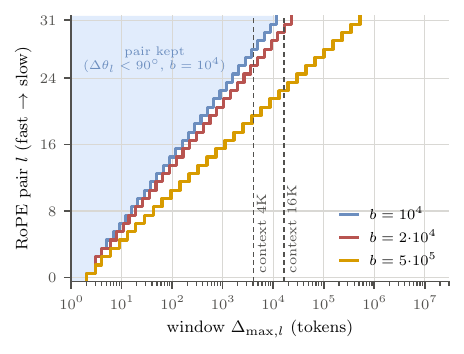}
  \caption{Which score parts \method{} keeps ($\Phimax = 90^\circ$, $d = 64$). Pair $l$ contributes
    only while its RoPE rotation $\Delta\theta_l$ is below the angle budget, i.e.\ up to the distance
    $\Delta_{\max,l} = \Phimax / \theta_l$ (shaded for $b = 10^4$). Fast pairs see a few tokens, slow pairs thousands; raising
    the RoPE base $b$ stretches every window.}
  \label{fig:windows}
\end{figure}

\subsection{The \method{} rule}

\method{} defines sparsity using a simple criterion: the score part of a pair $s_{ij}^{(l)}$ is kept only while the rotation induced by RoPE between the query and the key stays below a fixed angle. The projections that produce the queries $\mathbf{q}$ and the keys $\mathbf{k}$ are left untouched and only the dot product between them becomes partial.

We keep the notation of \cref{eq:decomp}. A head of dimension $d$ consists of $P = d/2$ RoPE pairs, ordered from the
fastest ($l = 0$) to the slowest ($l = P-1$). Pair $l$ of the query at position $i$ and of the key at position $j$ is
denoted $\pairq{l}_i$ and $\pairk{l}_j$ before rotation, and rotates at the frequency $\theta_l = b^{-2l/d}$ for a RoPE
base $b$. The distance between a query and a key is $\Delta = i - j \ge 0$ and a pair $l$ is thus rotated by
$\Delta\theta_l$. Given an \emph{angle budget} $\Phimax$, \method{} keeps only the pairs whose rotation is still within
the budget:
\begin{equation}
  \boxed{\;
  s^{\text{\method{}}}_{ij} = \frac{1}{\sqrt{d}} \sum_{l=0}^{P-1} \mathbf{1}\big[\Delta\,\theta_l < \Phimax\big]\,
  \big\langle R_{i\theta_l}\pairq{l}_i,\, R_{j\theta_l}\pairk{l}_j \big\rangle\;}
  \label{eq:hisa}
\end{equation}
where $R_{\alpha}$ denotes the rotation matrix by an angle $\alpha$ from RoPE and $\mathbf{1}[\cdot]$ the indicator function. Each
summand is exactly the corresponding score part $s^{(l)}_{ij}$ of the dense score (\cref{eq:part}): the indicator is
the only change, and it removes a score part as soon as its rotation exceeds the budget. With the default budget $\Phimax = 90^\circ$, a pair
contributes as long as RoPE has turned it by less than a quarter of a turn.

Since the criterion depends on the distance alone, it gives each pair its own \emph{window}: pair $l$ contributes to
keys at most $\Delta_{\max,l} = \Phimax/\theta_l$ positions away and is ignored beyond. The rotation speed decreases with
$l$ and so the windows grow. \cref{fig:windows} shows that the fastest pairs see only a few tokens and the
slowest several thousand. A nearby key is therefore compared with the query on all pairs, a more distant one only on the slower pairs. A key leaves the softmax entirely only once no pair is active, that is, beyond the \emph{reach}
\begin{equation}
  R = \frac{\Phimax}{\theta_{P-1}} = \Phimax\, b^{(d-2)/d},
  \label{eq:reach}
\end{equation}
set by the slowest pair. The attention output is then computed as usual, $\mathbf{o}_i = \sum_j \softmax_j(s^{\text{\method{}}}_{ij})\,\mathbf{v}_j$.

\paragraph{Static, parameter-free sparsity.}
Token-sparse attention removes entries of the $T \times T$ score matrix, while \method{} keeps the entries and removes, for each of them, some of the $P$ score parts that make up the score. The pattern is static: it depends only on the distance $i - j$ and on the RoPE frequencies, not on the content of the queries and keys. \method{} therefore adds no parameter, no router and no top-$k$ selection; the only new quantity is the hyperparameter $\Phimax$, and the projections, the values and the softmax are unchanged. Two consequences matter in practice. First, the model is trained with exactly the pattern it uses at inference. Second, which parts are needed is known before any token is seen, so kernels skip them without a selection pass and the KV cache drops each part of a key once it leaves the window (\cref{sec:exp-kv}). With $\Phimax = \infty$, \method{} is dense attention.

\paragraph{Angle budget and RoPE base.}
The angle budget $\Phimax$ and the RoPE base $b$ jointly fix every window: $\Delta_{\max,l} = \Phimax\, b^{2l/d}$. Raising $b$
stretches all windows and the reach (\cref{eq:reach}), which is how \method{} adapts to long contexts
(\cref{sec:exp-long}); lowering $\Phimax$ shortens them. For instance, with $d = 64$, $\Phimax = 90^\circ$ and
$b = 10^4$ the fastest pair sees 2 tokens and the slowest about 11.8K; with $b = 2\cdot 10^4$ the reach becomes about
23K tokens.

\paragraph{Compute and memory.}
The fraction of the dense causal dot product that \method{} computes over a sequence of length $T$ is
\begin{equation}
  \rho(T) = \frac{1}{P} \sum_{l=0}^{P-1} \frac{\sum_{\Delta=0}^{T-1} (T - \Delta)\, \mathbf{1}[\Delta < \Delta_{\max,l}]}
  {T(T+1)/2}.
  \label{eq:fraction}
\end{equation}
\cref{app:fraction} gives $\rho$ for other budgets, bases and lengths. During decoding, pair $l$ of a cached key can be dropped once the key is older than $\Delta_{\max,l}$, and the whole token once it is older than $R$.

\section{Experiments}
\label{sec:exp}

\paragraph{Setup.}
We train decoder-only models with the SmolLM-135M and SmolLM-360M architectures \citep{allal2024smollm} (head dimension $d = 64$, i.e.\ 32 RoPE pairs, grouped-query attention \citealp{ainslie2023gqa}) from scratch on OpenWebText \citep{gokaslan2019openwebtext}, for about 20 tokens per parameter, with AdamW, a cosine schedule and a peak learning rate selected for the dense model. Dense and \method{} runs share every setting and differ only in the attention rule. We write \textbf{\method{}} for the exact rule (\cref{eq:hisa}), computed by our Triton \citep{tillet2019triton} kernels, and \textbf{Flash-\method{}} for its FlashAttention-3 implementation, which uses a block-level version of the rule (\cref{sec:exp-fa}). \Cref{app:details} gives the full recipe and \cref{app:kernels} gives additional details on the kernels.

\subsection{\method{} vs.\ dense attention}
\label{sec:exp-4k}

\Cref{tab:main-4k} compares \method{} ($\Phimax = 90^\circ$) with dense attention at 4K context size. \method{} reaches a lower validation loss than the complete dot product while computing 29--34\% of the dense dot products, and the gap is the same at every position of the predicted token in the context window. It also holds at 360M parameters with a Chinchilla budget of 7.24B tokens.

\begin{table}[H]
  \caption{Dense attention vs.\ \method{} ($\Phimax = 90^\circ$) at 4K context: validation loss, loss by position, and
    percentage $\rho$ of the dot product computed. 135M: 2.7B tokens; 360M: 7.24B tokens. Best in bold.}
  \label{tab:main-4k}
  \centering
  \small
  \resizebox{\columnwidth}{!}{
\begin{tabular}{llrrrrr}
\toprule
Model & & Val.\ loss & \multicolumn{3}{c}{Loss by position} & $\rho$ (\%) \\
\cmidrule(lr){4-6}
 & & & 0--512 & 512--2K & 2K--4K & \\
\midrule
135M, $b=10^4$ & Dense & 2.906 & 3.170 & 2.920 & 2.830 & 100.0 \\
 & \method{} & \textbf{2.880} & \textbf{3.147} & \textbf{2.893} & \textbf{2.804} & 29.3 \\
\midrule
135M, $b=2{\cdot}10^4$ & Dense & 2.908 & 3.177 & 2.923 & 2.829 & 100.0 \\
 & \method{} & \textbf{2.869} & \textbf{3.136} & \textbf{2.883} & \textbf{2.791} & 34.2 \\
\midrule
360M, $b=10^4$ & Dense & 2.587 & 2.823 & 2.598 & 2.519 & 100.0 \\
 & \method{} & \textbf{2.575} & \textbf{2.810} & \textbf{2.585} & \textbf{2.508} & 29.3 \\
\bottomrule
\end{tabular}
}
\end{table}

The angle budget $\Phimax = 90^\circ$ and the other choices of this recipe come from the ablations of \cref{app:ablations}: every finite budget from $45^\circ$ to $360^\circ$ beats dense attention and $90^\circ$ is a compromise between compute and loss. \Cref{sec:interp} relates the dropped score parts to what a dense model learns.

\subsection{Flash-\method{}: \method{} in FlashAttention}
\label{sec:exp-fa}

\begin{figure}[t]
  \centering
  \includegraphics[width=\columnwidth]{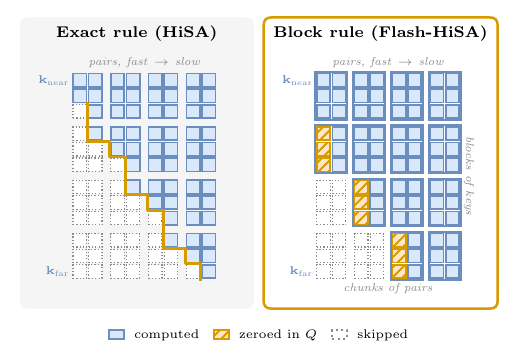}
  \caption{Exact rule and block rule, for one query (block). Rows: keys, nearest first; columns: RoPE pairs, fast to
    slow. The exact rule (left) drops pairs key by key. The block rule (right) groups keys into blocks (128 tokens) and
    pairs into chunks (8 pairs, 16 features); each block keeps the pairs of its nearest key, so a (block, chunk) tile
    of $K$ is either skipped or multiplied, with the dropped pairs of the boundary chunk zeroed in $Q$.}
  \label{fig:blockrule}
\end{figure}

When using the exact rule, each pair's window ends at a different distance, so a pair is kept for some entries of a FlashAttention tile and not for others which is not efficient for GPU's vectorized operations. Flash-\method{} follows the exact rule but applies it block by block instead. Queries and keys are cut along the sequence into blocks of $S = 128$ tokens, and the features into chunks of 16 (8 pairs, one matrix-multiply step on Hopper GPUs), so $K$ is a grid of $128 \times 16$ tiles. For a query block and a key block, the kept pairs are those that the exact rule keeps at the \emph{closest} distance between the two blocks. Each tile of $K$ is then multiplied in full or skipped, and the one chunk cut by the rule is multiplied with its dropped pairs set to zero in $Q$ (\cref{fig:blockrule}). This block rule keeps every score part of the exact rule plus up to $S-1$ tokens more per pair: an approximation of \method{} that fits the tiling of FlashAttention, and a different model (training and inference must use the same $S$). We implement it in FlashAttention-3 \citep{shah2024flashattention3}, forward and backward (\cref{app:kernels}).

\Cref{tab:fa} compares the implementations. The exact rule has the best loss but trains 1.7 times slower than dense
attention. Flash-\method{} computes slightly more of the dot product (35\% instead of 29\%), keeps about half of the gain
($-0.012$ against dense FlashAttention-3), and trains at the speed of dense FlashAttention-3.

\begin{table}[H]
  \caption{Implementations at 4K context ($b = 10^4$, $\Phimax = 90^\circ$, $S = 128$, $135$M parameters): validation loss, training
    throughput on one H200, and percentage $\rho$ of the dot product computed. Best in bold.}
  \label{tab:fa}
  \centering
  \small
  \resizebox{\columnwidth}{!}{
\begin{tabular}{llrrr}
\toprule
Attention & Kernel & Val.\ loss & Tokens/s & $\rho$ (\%) \\
\midrule
Dense & SDPA (FlashAtt.\ 2) & 2.906 & 205k & 100.0 \\
Dense & FlashAtt.\ 3 & 2.904 & 236k & 100.0 \\
\method{} & Triton (exact rule) & \textbf{2.880} & 119k & 29.3 \\
\method{} block rule & Triton & 2.888 & 191k & 34.8 \\
Flash-\method{} & FlashAtt.\ 3 & 2.892 & \textbf{238k} & 34.8 \\
\bottomrule
\end{tabular}
}
\end{table}

\subsection{Long-context extension}
\label{sec:exp-long}

We pretrain dense and \method{} models at 4K context size with a RoPE base $b = 2\cdot10^4$, which gives \method{} a reach of 23K tokens, then
extend them to 16K with 540M tokens (20\% of pretraining) of PG19 books \citep{rae2019compressive} and OpenWebText,
following \citet{xiong2023effective,gao2024prolong} (details in \cref{app:details}). After the extension, \method{}
keeps its advantage on books and web text while computing a fifth of the dot product (\cref{tab:long-exact}).

\begin{table}[H]
  \caption{After the 16K extension ($b = 2\cdot10^4$), exact rule: validation loss on PG19 (overall and by position) and
    OpenWebText, and percentage $\rho$ of the dot product computed. Best in bold.}
  \label{tab:long-exact}
  \centering
  \small
  {\footnotesize\setlength{\tabcolsep}{2.5pt}
\begin{tabular}{lrrrrrr}
\toprule
 & \multicolumn{4}{c}{PG19 val.\ loss} & OWT & $\rho$ (\%) \\
\cmidrule(lr){2-5}
 & all & 0--4K & 4K--8K & 8K--16K & val.\ loss & \\
\midrule
Dense & 3.205 & 3.262 & 3.209 & 3.174 & 3.111 & 100.0 \\
\method{} & \textbf{3.170} & \textbf{3.226} & \textbf{3.174} & \textbf{3.140} & \textbf{3.071} & 20.2 \\
\bottomrule
\end{tabular}
}
\end{table}

Flash-\method{}, pretrained and extended with the same recipe, also keeps its advantage, with a smaller gap
(\cref{tab:long-fa}). It trains at 94\% of the speed of dense FlashAttention-3 although it computes only 22\% of the dot
product.

\begin{table}[H]
  \caption{After the 16K extension with FlashAttention-3 ($b = 2\cdot10^4$): validation loss, percentage $\rho$ of the
    dot product computed, and training throughput on one H200. Best in bold.}
  \label{tab:long-fa}
  \centering
  \small
  \resizebox{\columnwidth}{!}{\footnotesize\setlength{\tabcolsep}{2.5pt}
\begin{tabular}{lrrrrrrr}
\toprule
 & \multicolumn{4}{c}{PG19 val.\ loss} & OWT & $\rho$ (\%) & Tok/s \\
\cmidrule(lr){2-5}
 & all & 0--4K & 4K--8K & 8K--16K & val.\ loss & & \\
\midrule
Dense (FA3) & 3.212 & 3.270 & 3.216 & 3.181 & 3.115 & 100.0 & \textbf{143k} \\
Flash-\method{} & \textbf{3.190} & \textbf{3.248} & \textbf{3.194} & \textbf{3.159} & \textbf{3.099} & 21.7 & 135k \\
\bottomrule
\end{tabular}
}
\end{table}

\subsection{KV cache and decoding}
\label{sec:exp-kv}

A cached key needs pair $l$ only for $\Delta_{\max,l}$ tokens and its value only within the reach. Flash-\method{} stores each 16-feature key chunk in its own ring buffer, as long as the window of its slowest pair, and its decoding kernel reads only the chunks still in use (\cref{app:kernels}). The cache is a third smaller (\cref{tab:decode}) than the dense one at 4K and 16K, and decoding is faster than dense FlashAttention-3, increasingly so with the context: $+42\%$ tokens per second at 16K and batch 128. Part of this comes from our kernel's smaller query tiles; with the same kernel and no pair dropped, Flash-\method{} is still 13\% faster.

\begin{table}[H]
  \caption{Decoding with FlashAttention-3 at full context: throughput (tokens/s, one H200) and KV-cache memory, for
    batch sizes 32 and 128. 4K: $b = 10^4$; 16K: $b = 2\cdot10^4$, after the extension. Best in bold.}
  \label{tab:decode}
  \centering
  \small
  \resizebox{\columnwidth}{!}{
\begin{tabular}{llrrrr}
\toprule
 & & \multicolumn{2}{c}{Batch 32} & \multicolumn{2}{c}{Batch 128} \\
\cmidrule(lr){3-4}\cmidrule(lr){5-6}
$T$ & Attention & Tok/s & KV (GiB) & Tok/s & KV (GiB) \\
\midrule
4K & Dense (FA3) & 7.3k & 2.8 & 20.4k & 11.2 \\
 & Flash-\method{} & \textbf{7.8k} & \textbf{1.9} & \textbf{23.2k} & \textbf{7.7} \\
\midrule
16K & Dense (FA3) & 4.7k & 11.2 & 9.2k & 45.0 \\
 & Flash-\method{} & \textbf{5.6k} & \textbf{7.3} & \textbf{13.0k} & \textbf{29.1} \\
\bottomrule
\end{tabular}
}
\end{table}

\subsection{Converting a trained dense model}
\label{sec:exp-convert}

As existing models rely mostly on dense attention, we also investigate whether a model pre-trained with dense attention can be converted to \method{} and match the performance of dense attention under the same training budget. To test this scenario, we take the dense 135M model of \cref{tab:main-4k} ($b = 10^4$, 4K context), switch its attention to \method{} without changing any weight, and continue training for 1.35B tokens. A dense control continues with the same recipe. After continued training (\cref{tab:convert}), the converted model matches the dense one while computing 29\% of the dot products.

\begin{table}[H]
  \caption{Converting the dense 135M model ($b = 10^4$, 4K context) to \method{}: validation loss before and after
    continued training, and percentage $\rho$ of the dot product computed. Best in bold.}
  \label{tab:convert}
  \centering
  \small
  \resizebox{\columnwidth}{!}{
\begin{tabular}{lrrr}
\toprule
Model & Extra tokens & Val.\ loss & $\rho$ (\%) \\
\midrule
Dense (start) & -- & 2.906 & 100.0 \\
Dense $\rightarrow$ \method{}, no training & 0 & 2.937 & 29.3 \\
Dense, continued & 1.35B & 2.838 & 100.0 \\
Dense $\rightarrow$ \method{}, continued & 1.35B & \textbf{2.836} & 29.3 \\
\bottomrule
\end{tabular}
}
\end{table}

\section{Dense attention learns to suppress the score parts \method{} drops}
\label{sec:interp}

To provide an interpretability perspective on our work, we test whether the attention score parts $s^{(l)}_{ij}$ are being used or ignored by dense models when $\Delta\theta_l$ is large. For every frequency pair and distance, we measure how much the pair's dot product moves the attention output compared with an uninformative score part of the same rotation. Its effect can be above this noise level, at it, or below it, which indicates that the model actively suppresses it. We follow this measure in our dense 4K model during training and in the pretrained SmolLM-135M.

\paragraph{Measure.}
For a head $h$, a pair $l$, and a band of distances $\mathcal{B}$, let $s^{(l)}_{ij}$ be the attention score part of pair $l$ (\cref{eq:part}, pre-softmax) with query pair
$\pairq{l}_i$ and key pair $\pairk{l}_j$ and $\mathbf{v}_j$ the values. The bands split the query--key distance $i - j$ on a logarithmic scale: $\{0\}$,
$\{1\}$, $[2, 4)$, $[4, 8)$, \dots, $[1024, 2048)$, twelve bands in total. A band thus selects, for every query, the keys
at a given range of distances, and the measure tells how much pair $l$ matters at that range. 
For a query $i$, the head output is the attention-weighted average of the values,
\begin{equation}
  \mathbf{o}_i = \sum_j a_{ij}\,\mathbf{v}_j, \qquad a_{ij} = \frac{e^{s_{ij}}}{\sum_{k} e^{s_{ik}}},
\end{equation}
with $s_{ij} = \sum_l s^{(l)}_{ij}$ the attention score of \cref{eq:part}. To measure what pair $l$ contributes at the distances of a band $\mathcal{B}$, we compute the same output without its
score part for the keys in that band:
\begin{equation}
  s'_{ij} =
  \begin{cases}
    s_{ij} - s^{(l)}_{ij} & \text{if } i - j \in \mathcal{B}, \\
    s_{ij}                & \text{otherwise},
  \end{cases}
\end{equation}
and
\begin{equation}
  \mathbf{o}'_i = \sum_j a'_{ij}\,\mathbf{v}_j, \qquad a'_{ij} = \frac{e^{s'_{ij}}}{\sum_{k} e^{s'_{ik}}}.
\end{equation}

The quantity of interest is the change of the output,
\begin{equation}
  \Delta\mathbf{o}_i = \mathbf{o}'_i - \mathbf{o}_i
  \label{eq:delta-o}
\end{equation}
(it depends on $l$ and $\mathcal{B}$, as do $s'_{ij}$, $a'_{ij}$ and $\mathbf{o}'_i$; we omit them from the notation for clarity),
which can be written as (derivation in \cref{app:interp})
\begin{equation}
  \Delta\mathbf{o}_i = \frac{\sum_j u_j\,(\mathbf{v}_j - \mathbf{o}_i)}{1 + \sum_j u_j},
  \label{eq:delta-o2}
\end{equation}
where
\begin{equation}
  u_j = a_{ij}\big(e^{-s^{(l)}_{ij}} - 1\big)\,\mathbf{1}[i - j \in \mathcal{B}]
  \label{eq:w}
\end{equation}
is the change of the attention weight of key $j$ before renormalization.

To measure this change at the output of the attention layer rather than inside each head, we map it through
$W_h$, the output projection of head $h$. As a reference, we compute the same change with $s^{(l)}_{ij}$ replaced by
$\hat s^{(l)}_{ij}$, the score part of pair $l$ computed with pair $l$ of the key of \emph{another} document at the
same position (same rotation, unrelated content), while $a_{ij}$, $\mathbf{v}_j$ and $\mathbf{o}_i$ stay those of the
true document; we write $\Delta\hat{\mathbf{o}}_{h,i}$ for this change.

Both changes are computed on $N = 256$ OpenWebText validation documents of at least 2048 tokens: the model reads the
first 2048 tokens of each document, and every token serves as a query (details in \cref{app:interp}). We accumulate
the squared changes over the documents, the heads of the layer and all queries,
\begin{equation}
  \begin{aligned}
    E^{\text{real}}_{l,\mathcal{B}} &= \sum_{n=1}^{N} \sum_{h} \sum_{i} \big\| W_h \, \Delta\mathbf{o}^{(n)}_{h,i} \big\|^2, \\
    E^{\text{null}}_{l,\mathcal{B}} &= \sum_{n=1}^{N} \sum_{h} \sum_{i} \big\| W_h \, \Delta\hat{\mathbf{o}}^{(n)}_{h,i} \big\|^2,
  \end{aligned}
  \label{eq:e-real}
\end{equation}
where $\Delta\mathbf{o}^{(n)}_{h,i}$ is \cref{eq:delta-o} for head $h$ and document $n$; $E^{\text{null}}_{l,\mathcal{B}}$ is
the same sum with $\Delta\hat{\mathbf{o}}^{(n)}_{h,i}$. We report the ratio $E^{\text{real}}_{l,\mathcal{B}} / E^{\text{null}}_{l,\mathcal{B}}$.

A ratio above 1 means that the true score part moves the output more than an uninformative one (content-specific signal);
below 1, it moves the output \emph{less} than an unrelated key would, so the model has shaped its queries and keys to
damp it.

\begin{figure}[t]
  \centering
  \includegraphics[width=\columnwidth]{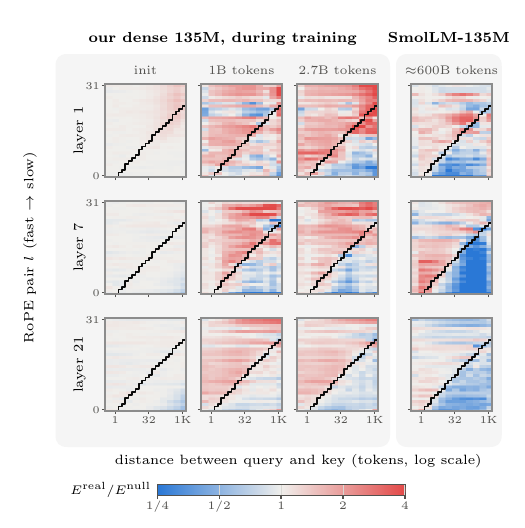}
  \caption{$E^{\text{real}} / E^{\text{null}}$ in three layers of the dense 135M model at initialisation, after 1B tokens
    and at the end of training (2.7B tokens), and of the pretrained SmolLM-135M. In each panel, rows are frequency
    pairs (fast at the bottom) and columns distance bands (1 to 2048 tokens, log scale). Red: the true score part moves the
    head output more than one computed with an unrelated key; blue: less. Black line: $\Delta\theta_l = 90^\circ$
    (right of it, \method{} drops the score part). All layers in \cref{app:interp}.}
  \label{fig:interp-train}
\end{figure}

\begin{figure}[t]
  \centering
  \includegraphics[width=\columnwidth]{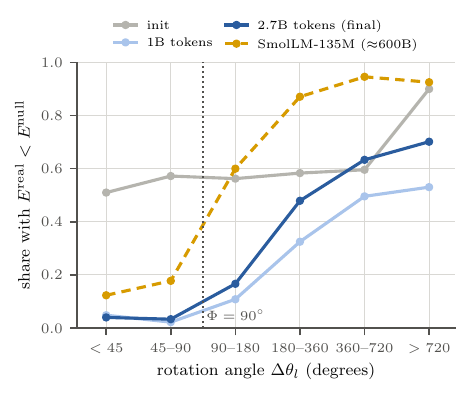}
  \caption{Share of the (layer, distance band) cells of the fast pairs ($l = 0, \dots, 7$) below the uninformative
    reference ($E^{\text{real}} < E^{\text{null}}$), by rotation angle, at initialisation, after 1B tokens and at the
    end of training of the dense 135M model, and for the pretrained SmolLM-135M (all 30 layers).}
  \label{fig:interp-share}
\end{figure}

\paragraph{Results.}

At initialization, the heatmaps are uniformly close to 1 (\cref{fig:interp-train}) and start to shift during training. During the first steps the model learns to use the content of the pair scores at small rotation angles, which turn red, left of the $90^\circ$ line. Right of the line, the scores are slowly turning blue. While this requires more training steps, some layers suppress scores at large distances aggressively. This is particularly visible in SmolLM-135M (all heatmaps and details are in \cref{app:interp}). 

\Cref{fig:interp-share} summarises all layers for the fast pairs ($l = 0, \dots, 7$). At initialisation, most cells lie at the reference. The model quickly learns to use the scores at small angles: after 1B tokens, only 2--5\% of the cells below $90^\circ$ are under the reference. Suppressing the scores at large angles takes much longer: their share under the reference grows slowly over training (e.g.\ from 33\% to 48\% at $180$--$360^\circ$) and reaches 87--95\% beyond $180^\circ$ in SmolLM-135M, trained on about 600B tokens.

Their relative influence on the head outputs, which needs no reference, decreases accordingly: beyond
$720^\circ$ it falls from 4.6\% of the influence of the parts below $45^\circ$ after 1B tokens to 2.1\% at the end of
training and 0.5\% in SmolLM-135M (\cref{app:interp}).

A dense model thus spends part of its training suppressing the score parts with a large rotation, and this suppression is still progressing at the end of our runs. \method{} builds it into the architecture, which is consistent with its lower loss and with the small cost of converting a dense model (\cref{sec:exp-convert}).

\section{Conclusion}
\label{sec:conclusion}

We asked whether attention needs the complete query--key dot product. With RoPE, the score splits into one part per frequency pair, and \method{} keeps each part only while its rotation stays within an angle budget. This sparsity needs no learned selection: it is fixed by the positions, so it is known before any token is seen. In our preliminary experiments with 135M- and 360M-parameter models, this partial product gives a lower validation loss than the complete one while computing only part of it. We also developed Flash-\method{}, a FlashAttention-3 implementation of \method{}, which trains at a similar speed to dense FlashAttention-3 while increasing decoding speed and reducing memory footprint during inference. The advantage persists when using long-context extension and a model trained with dense attention can be converted to \method{} by continued training without losing quality. An analysis of the head outputs suggests why: during training, a dense model learns to suppress the score pairs with large rotations, which \method{} removes by construction. Open questions such as scaling to bigger models, downstream tasks benchmarking and better kernel design are left for future work.

\section*{Impact Statement}
This paper presents work whose goal is to advance the field of Machine Learning, in particular the efficiency of
attention in large language models. Cheaper training and inference lower the cost and energy of running such
models; we see no societal consequences specific to this method beyond those of language models in general.

\bibliography{references}
\bibliographystyle{icml2026}

\newpage
\appendix
\crefalias{section}{appendix}       
\crefalias{subsection}{subappendix}
\onecolumn

\section{Compute and memory of \method{}}
\label{app:fraction}

\paragraph{Fraction of the dot product computed.}
With integer windows $L_l = \lceil \Delta_{\max,l} \rceil$ (pair $l$ is kept iff $\Delta < L_l$), pair $l$ contributes to
$m_l T - m_l(m_l - 1)/2$ of the $T(T+1)/2$ causal query--key entries, with $m_l = \min(L_l, T)$, which gives
$\rho(T)$ of \cref{eq:fraction} (\cref{tab:fraction}). The windows grow with the pair index,
$\Delta_{\max,l} = \Phimax\, b^{2l/d}$, so pairs whose window exceeds $T$ contribute fully and the others contribute little: $\rho$
is close to the share of pairs with $\Delta_{\max,l} \gtrsim T$, which falls by $d/(2\log_2 b)$ pairs, about 7.5 percentage points
at $b = 10^4$, per doubling of $T$ (37\%, 29\% and 14\% at 2K, 4K and 16K). Doubling the budget or the base has the
same effect in the other direction. The block rule of Flash-\method{} (\cref{fig:blockrule}) uses the minimum distance
between blocks instead of $\Delta$ and computes a bit more (35\% at 4K and 16\% at 16K, instead of 29\% and 14\%).

\paragraph{Training versus decoding.}
$\rho(T)$ counts the entries of the causal score matrix, where distance $\Delta$ occurs $T - \Delta$ times: it is the
fraction computed in a full forward pass (training, prefill), dominated by short distances. One decoding step at
context $t$ compares the new query with $t$ keys, one at each distance $0, \dots, t-1$, so the fraction it computes is
\begin{equation}
  \rho_{\text{dec}}(t) = \frac{1}{P} \sum_{l} \frac{\min(L_l, t)}{t},
\end{equation}
which weighs all distances equally and is smaller than $\rho(t)$ (\cref{tab:fraction-decode}). The two are consistent:
a full pass is the sum of its decoding steps, $\rho(T) = \sum_{t=1}^{T} t\, \rho_{\text{dec}}(t) \,/\, \sum_{t=1}^{T} t$.

\paragraph{Which work is removed.}
$\rho$ counts the multiply--adds of the query--key product. The other parts of attention, the softmax and the product
with the values (and, in training, most of the backward pass), scale with the number of query--key entries, which
stays the same as long as every key is within the reach. They only shrink for keys beyond the reach, i.e.\ for a
fraction $(T - R)(T - R + 1) / (T(T+1))$ of the entries when $T > R$. This is why the compute saving appears in
decoding, where reading the cache dominates, before it appears in training (\cref{sec:exp-long}).

\paragraph{KV-cache memory.}
During decoding at context $T$, pair $l$ of a cached key is needed for the last $\min(L_l, T)$ tokens and the value
for the last $\min(R, T)$. Keys and values have the same size, so the cache, relative to a dense one, is
\begin{equation}
  \kappa(T) = \frac{1}{2}\Big(\frac{1}{P} \sum_{l} \frac{\min(L_l, T)}{T} + \frac{\min(R, T)}{T}\Big).
\end{equation}
While $T \le R$ every value is kept, so the saving is at most 50\%; beyond the reach, keys and values are both evicted.
Our caches store keys per 16-feature chunk, with a ring as long as the window of the chunk's slowest pair (rounded up to
128-token blocks for Flash-\method{}), which keeps slightly more than per-pair storage: 66.5\% of a dense cache for
\method{} and 68.4\% for Flash-\method{} at 4K ($b = 10^4$), against 62\% per pair; 64.6\% for Flash-\method{} at 16K
($b = 2\cdot10^4$).

\begin{table}[h]
  \caption{Percentage of the dense query--key dot product computed by \method{}, $\rho(T)$ (\cref{eq:fraction}), for
    angle budgets $\Phimax$, RoPE bases $b$ and context lengths $T$ ($d = 64$, causal attention), and the reach $R$
    at $90^\circ$ (\cref{eq:reach}).}
  \label{tab:fraction}
  \centering
  \small
\begin{tabular}{lrrrrrrrrrrrrr}
\toprule
$b$ & \multicolumn{4}{c}{$T={}$2K} & \multicolumn{4}{c}{$T={}$4K} & \multicolumn{4}{c}{$T={}$16K} & reach \\
\cmidrule(lr){2-5}\cmidrule(lr){6-9}\cmidrule(lr){10-13}
 & $45^\circ$ & $90^\circ$ & $180^\circ$ & $360^\circ$ & $45^\circ$ & $90^\circ$ & $180^\circ$ & $360^\circ$ & $45^\circ$ & $90^\circ$ & $180^\circ$ & $360^\circ$ & ($90^\circ$) \\
\midrule
$10^4$ & 29.3 & 36.9 & 44.4 & 51.9 & 21.8 & 29.3 & 36.8 & 44.4 & 8.1 & 14.3 & 21.8 & 29.3 & 11.8K \\
$2{\cdot}10^4$ & 34.2 & 41.2 & 48.2 & 55.1 & 27.2 & 34.2 & 41.2 & 48.1 & 13.2 & 20.2 & 27.2 & 34.2 & 23.1K \\
$5{\cdot}10^4$ & 39.6 & 46.0 & 52.4 & 58.8 & 33.2 & 39.6 & 46.0 & 52.4 & 20.4 & 26.8 & 33.2 & 39.6 & 56.0K \\
$5{\cdot}10^5$ & 50.0 & 55.2 & 60.5 & 65.8 & 44.6 & 49.9 & 55.2 & 60.5 & 34.1 & 39.4 & 44.6 & 49.9 & 521.2K \\
\bottomrule
\end{tabular}

\end{table}

\begin{table}[h]
  \caption{Percentage of the dense query--key product computed in a full causal pass, $\rho(T)$, and in one decoding
    step at context $T$, $\rho_{\text{dec}}(T)$, and KV-cache size relative to a dense cache, $\kappa(T)$ (per pair),
    at $\Phimax = 90^\circ$ ($d = 64$).}
  \label{tab:fraction-decode}
  \centering
  \small
\begin{tabular}{lrrrrrrrrr}
\toprule
$b$ & \multicolumn{3}{c}{$T={}$2K} & \multicolumn{3}{c}{$T={}$4K} & \multicolumn{3}{c}{$T={}$16K} \\
\cmidrule(lr){2-4}\cmidrule(lr){5-7}\cmidrule(lr){8-10}
 & $\rho$ & $\rho_{\text{dec}}$ & $\kappa$ & $\rho$ & $\rho_{\text{dec}}$ & $\kappa$ & $\rho$ & $\rho_{\text{dec}}$ & $\kappa$ \\
\midrule
$10^4$ & 36.9 & 31.5 & 65.7 & 29.3 & 23.9 & 61.9 & 14.3 & 9.0 & 40.4 \\
$2{\cdot}10^4$ & 41.2 & 36.1 & 68.1 & 34.2 & 29.1 & 64.5 & 20.2 & 15.1 & 57.6 \\
$5{\cdot}10^4$ & 46.0 & 41.4 & 70.7 & 39.6 & 35.0 & 67.5 & 26.8 & 22.1 & 61.1 \\
$5{\cdot}10^5$ & 55.2 & 51.4 & 75.7 & 49.9 & 46.1 & 73.1 & 39.4 & 35.5 & 67.8 \\
\bottomrule
\end{tabular}

\end{table}

\section{Experimental details}
\label{app:details}

\paragraph{Models.}
\Cref{tab:arch} lists the two architectures, which follow SmolLM \citep{allal2024smollm}: LLaMA-style decoder blocks with
RMSNorm, SwiGLU MLPs, tied input and output embeddings (vocabulary of 49{,}152 tokens) and RoPE with base $10^4$ unless
stated otherwise. All weights are initialised randomly; only the architecture is taken from SmolLM.

\begin{table}[h]
  \caption{Model architectures and training budgets (about 20 tokens per parameter).}
  \label{tab:arch}
  \centering
  \small
\begin{tabular}{lrrrrrrr}
\toprule
Model & Params & Layers & Width & MLP & Heads (q / kv) & $d$ & Tokens \\
\midrule
135M & 134.5M & 30 & 576 & 1536 & 9 / 3 & 64 & 2.7B \\
360M & 361.8M & 32 & 960 & 2560 & 15 / 5 & 64 & 7.24B \\
\bottomrule
\end{tabular}

\end{table}

\paragraph{Optimisation.}
\Cref{tab:hparams} gives the training recipe, shared by every dense and \method{} run. The peak learning rate is selected
for the dense model only, with short runs at several values (\cref{tab:lr}); we keep the value with the lowest validation
loss, or the lower of the two best when they are within 0.02.

\begin{table}[h]
  \caption{Training hyperparameters.}
  \label{tab:hparams}
  \centering
  \small
  \begin{tabular}{ll}
    \toprule
    Optimizer & AdamW, $\beta = (0.9, 0.95)$, weight decay 0.1 on matrices, gradient clipping 1.0 \\
    Batch & 524{,}288 tokens (128 sequences of 4K, 256 of 2K, 32 of 16K) \\
    Schedule & linear warmup over 1\% of the steps (2\% for the 16K extension), cosine decay to 10\% of the peak \\
    Peak learning rate & $2.4\cdot10^{-3}$ (135M), $1.2\cdot10^{-3}$ (360M), $8\cdot10^{-4}$ (16K extension) \\
    Precision & bf16 autocast, fp32 master weights, \texttt{torch.compile} \\
    Seeds & 1337 (plus 1338 and 1339 for the 2K seed study) \\
    Hardware & one NVIDIA H200 GPU per run \\
    \bottomrule
  \end{tabular}
\end{table}

\begin{table}[h]
  \caption{Learning-rate selection on the dense model: validation loss after the given number of tokens, with the same
    number of warmup steps as the full run. For the 16K extension, the mean of the PG19 and OpenWebText validation
    losses.}
  \label{tab:lr}
  \centering
  \small
\begin{tabular}{llp{9.3cm}}
\toprule
Setting & Tokens & Peak LR: validation loss (selected in bold) \\
\midrule
135M, 2K, packed & 300M & 6.0e-4: 4.782, 1.2e-3: 4.718, \textbf{2.4e-3: 4.464} \\
135M, 4K, documents & 300M & 1.2e-3: 4.666, \textbf{2.4e-3: 4.430}, 3.6e-3: 4.622, 4.8e-3: 4.745 \\
360M, 4K, documents & 300M & 6.0e-4: 4.140, \textbf{1.2e-3: 3.958}, 2.4e-3: 3.949, 3.6e-3: 4.143 \\
135M, 16K extension & 100M & 2.4e-4: 3.404, \textbf{8.0e-4: 3.359}, 2.4e-3: 3.373 \\
\bottomrule
\end{tabular}

\end{table}

\paragraph{Data.}
OpenWebText is tokenized with the SmolLM tokenizer; 0.5\% of the documents (40{,}069) are held out for validation.
The 2K runs cut the token stream (documents separated by an end-of-sequence token) into windows of 2{,}049 tokens in a
seeded random order. The 4K runs use only windows that lie inside a single document of at least 4{,}097 tokens, cut
at a random offset at every epoch, so that the context is never filled with unrelated documents. The 16K extension
mixes, in fixed 50/50 proportions, single-book windows of PG19 \citep{rae2019compressive} (28{,}602 training books,
3.0B tokens) and packed OpenWebText windows. The continued-training runs (dense $\rightarrow$ \method{}) use the 4K
recipe with a new data seed.

\paragraph{Evaluation.}
The validation loss is the mean next-token cross-entropy over a fixed set of validation windows (3{,}200 windows at
2K, 1{,}600 at 4K, and 200 per source at 16K), also reported by position inside the window. Training throughput is the
mean over the last 200 steps of the run. Decoding is measured on the first $B$ validation windows: one token at a time
with teacher forcing, each step (all layers) captured as a CUDA graph, the KV cache allocated for the full context;
throughput is $B \cdot T$ divided by the time of the $T$ steps, and the decoded loss is checked against the loss of the
full forward pass.

\section{Choosing the recipe: angle budget, RoPE base, seeds and score normalisation}
\label{app:ablations}

These ablations led to the recipe of the main text ($\Phimax = 90^\circ$, windows from the model's own RoPE base).

\begin{table}[h]
  \caption{Angle budget $\Phimax$ at 2K context ($b = 10^4$, 135M): validation loss, percentage $\rho$ of the dot
    product computed, and reach. Best in bold.}
  \label{tab:abl-angle}
  \centering
  \small
\begin{tabular}{lrrr}
\toprule
$\Phimax$ & Val.\ loss & $\rho$ (\%) & Reach \\
\midrule
$45^\circ$ & 3.022 & 29.3 & 5.9K \\
$90^\circ$ & 3.020 & 36.9 & 11.8K \\
$180^\circ$ & \textbf{3.010} & 44.4 & 23.6K \\
$360^\circ$ & 3.039 & 51.9 & 47.1K \\
$\infty$ (dense) & 3.055 & 100.0 & -- \\
\bottomrule
\end{tabular}

\end{table}

\paragraph{Angle budget.}
\Cref{tab:abl-angle} varies $\Phimax$ at 2K context (one seed per budget). Every finite budget beats dense attention,
even $45^\circ$, which computes under a third of the dot product. $45^\circ$, $90^\circ$ and $180^\circ$ are within
0.012 of each other, about the spread between seeds (below), while a full rotation period ($360^\circ$, the window of
\citealp{du2026waveprune}) keeps less of the gain. The data do not single out one budget; we use $90^\circ$ as a
compromise between compute and loss. Removing the score parts with the largest rotation helps, consistent with the
analysis of \cref{sec:interp}, where a dense model learns to suppress them.

\paragraph{Seeds.}
At 2K context (135M, $b = 10^4$, $\Phimax = 90^\circ$), three seeds give $3.024 \pm 0.004$ for \method{} and
$3.052 \pm 0.011$ for dense attention; \method{} is better on each seed.

\paragraph{RoPE base.}
\Cref{tab:abl-base} pairs each \method{} model with a dense model at the same base. The gain holds at bases $10^4$ and
$2\cdot10^4$. With a rotation base of $5\cdot10^5$ but windows still computed at base $10^4$ (${}^\dagger$), the dense
model itself improves and the gap shrinks.

\begin{table}[h]
  \caption{RoPE base $b$ ($\Phimax = 90^\circ$, 135M): validation loss of dense attention and \method{} at the same
    base, and reach. ${}^\dagger$Rotation base $5\cdot10^5$, \method{} windows computed with base $10^4$. Best in
    bold.}
  \label{tab:abl-base}
  \centering
  \small
\begin{tabular}{llrrr}
\toprule
$T$ & $b$ & Dense & \method{} & Reach \\
\midrule
2K & $10^4$ & 3.055 & \textbf{3.020} & 11.8K \\
2K & $5{\cdot}10^5$\,$^\dagger$ & 3.019 & \textbf{3.010} & 11.8K \\
4K & $10^4$ & 2.906 & \textbf{2.880} & 11.8K \\
4K & $2{\cdot}10^4$ & 2.908 & \textbf{2.869} & 23.1K \\
\bottomrule
\end{tabular}

\end{table}

\paragraph{Score normalisation.}
\method{} keeps the $1/\sqrt{d}$ scale of \cref{eq:hisa} while dropping score parts, so the score of a distant key is
a sum over fewer pairs and its typical magnitude shrinks with the distance. The gain over dense attention could
therefore come from this implicit, content-independent down-weighting of distant keys rather than from removing the
score parts themselves. To test this, we train a variant that normalises each score by the number of pairs it keeps,
\begin{equation}
  s_{ij} = \frac{1}{\sqrt{2 n_{ij}}} \sum_{l=0}^{P-1} \mathbf{1}\big[\Delta\,\theta_l < \Phimax\big]\, \big\langle R_{i\theta_l}\pairq{l}_i,\, R_{j\theta_l}\pairk{l}_j \big\rangle,
  \qquad n_{ij} = \sum_{l=0}^{P-1} \mathbf{1}\big[\Delta\,\theta_l < \Phimax\big],
\end{equation}
so that every score has the scale of a dense dot product over its $2 n_{ij}$ features (4K context, 135M, $b = 10^4$,
one seed). It reaches a validation loss of 2.879, on par with \method{} (2.880) and well below dense attention
(2.906): the gain does not come from the shrinking scale of distant scores. The variant is however slower (87k
instead of 119k tokens/s with our Triton kernels, since the scale changes with every query--key distance) and worse on
a synthetic copy task at long range (repeating a 32-token span seen 2560 tokens earlier: 33\% token accuracy, against
77\% for \method{} and 60\% for dense attention). We therefore keep the $1/\sqrt{d}$ scale.

\section{Kernels}
\label{app:kernels}

\paragraph{Exact rule (Triton).}
The exact rule is computed by FlashAttention-2-style fused kernels for the forward and the backward pass, written in
Triton. The query and key features are permuted so that each 16-feature chunk holds 8 RoPE pairs (the two features of
each pair together). For a tile of query and key positions, the pairs whose window covers the whole tile, the pairs
that are dropped on the whole tile, and the ``mixed'' pairs whose window ends inside the tile are separated, and the
mixed pairs are masked element-wise; with integer thresholds $L_l = \lceil \Delta_{\max,l} \rceil$ the result equals the reference
rule exactly. At 4K context they train 1.7 times slower than dense attention with SDPA (\cref{tab:fa}).

\begin{figure}[h]
  \centering
  \includegraphics[width=\textwidth]{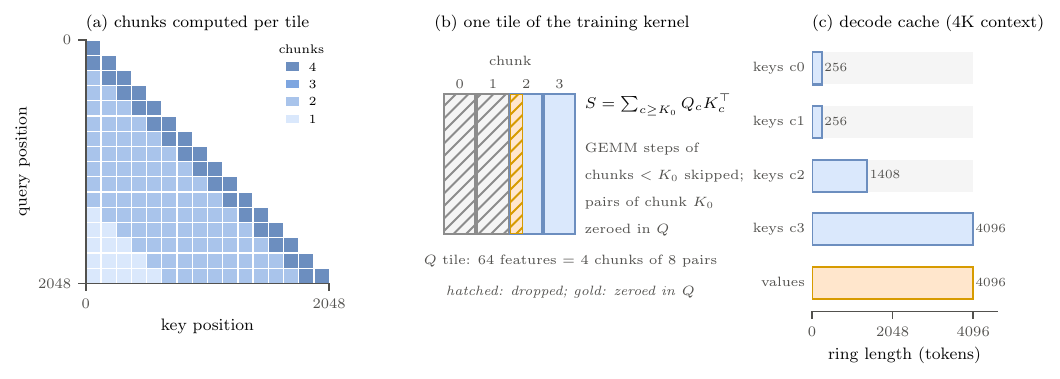}
  \caption{The FlashAttention implementation of \method{} (block rule, $S = 128$, $\Phimax = 90^\circ$, $b = 10^4$).
    (a) Number of 16-feature chunks computed for each $128 \times 128$ tile of the causal score matrix (2K context):
    near the diagonal all four, far from it only the slowest chunk. (b) One tile of the training kernel: the
    matrix multiply runs only over the kept chunks, and the dropped pairs of the boundary chunk are zeroed in the query
    tile. (c) The decoding cache at 4K context: each key chunk has its own ring buffer, as long as the window of its
    slowest pair; the values are kept within the reach.}
  \label{fig:kernel}
\end{figure}

\paragraph{Block rule in FlashAttention-3.}
With the block rule (\cref{fig:blockrule}) and a rule block equal to the key tile ($S = 128$), the set of kept pairs is the
same for every entry of a tile and depends only on the block distance (\cref{fig:kernel}a). Since windows grow with
the pair index, the dropped pairs are always the fastest ones: a tile at block distance $n$ drops the first $a_n$
pairs, i.e.\ the chunks below $K_0 = \lfloor a_n / 8 \rfloor$ and $a_n \bmod 8$ pairs of chunk $K_0$. We add this rule
to the FlashAttention-3 forward and backward kernels as a separate kernel instance. The query--key product of a tile
runs the tensor-core matrix multiply only over the chunks $K_0, \dots, 3$, and the dropped pairs of chunk $K_0$ are
zeroed in the shared-memory copy of the query tile (\cref{fig:kernel}b). The key blocks are visited from near to far,
so $K_0$ only grows along the loop: the loop is split into one straight segment per value of $K_0$, without a runtime
branch inside the asynchronous matrix-multiply pipeline, and the zeroing is cumulative. A small table (pairs dropped
per block distance, the four distances where a whole chunk drops, and the thresholds) is the only input that depends
on the angle budget and the RoPE base. Blocks beyond the reach are skipped entirely through the kernel's sliding-window
bounds. In the backward pass the recomputation of the scores skips the dropped chunks in the same way; the other
products (with $dO$, $V$ and the score gradients) still run at full width, which is why the backward pass does not yet
benefit from the sparsity (\cref{sec:exp-long}). Applying RoPE directly in the chunked feature layout avoids permuting
the queries and keys.

\paragraph{Decoding.}
During decoding there is one query per sequence. The query heads that share a KV head are packed as the rows of one
query tile of 64 rows (FlashAttention-3 uses 192 rows for $d = 64$, which would be almost entirely padding), and the
cache is split across thread blocks whose partial results are merged at the end. The keys are stored as four ring
buffers, one per 16-feature chunk, each as long as the window of its slowest pair rounded up to whole 128-token blocks,
and the values in a ring as long as the reach (\cref{fig:kernel}c). The token at position $j$ lives in slot
$j \bmod M_c$ of ring $c$, of length $M_c$; since every ring holds whole blocks, a key tile never wraps around. The producer threads load
the cache with asynchronous copies; each thread always loads the same chunk, and skips the load when its chunk is
dropped for the current key block, so the dropped chunks are never read from memory.

\newpage
\section{Interpretability: details and all layers}
\label{app:interp}

\paragraph{Derivation of \cref{eq:delta-o2}.}
Fix a query $i$, a pair $l$ and a band $\mathcal{B}$. Without the score part of pair $l$ on the band, the scores are
$s'_{ij} = s_{ij} - s^{(l)}_{ij}$ for $i - j \in \mathcal{B}$ and $s'_{ij} = s_{ij}$ otherwise, so
\begin{equation}
  e^{s'_{ij}} = e^{s_{ij}} \big(1 + (e^{-s^{(l)}_{ij}} - 1)\,\mathbf{1}[i - j \in \mathcal{B}]\big).
\end{equation}
Writing $e^{s_{ij}} = Z\, a_{ij}$ with $Z = \sum_k e^{s_{ik}}$ and using $u_j$ of \cref{eq:w}, the new normaliser and
the new weighted sum of the values are
\begin{equation}
  \sum_k e^{s'_{ik}} = Z \Big(1 + \sum_k u_k\Big), \qquad
  \sum_j e^{s'_{ij}} \mathbf{v}_j = Z \Big(\mathbf{o}_i + \sum_j u_j \mathbf{v}_j\Big),
\end{equation}
since $\sum_j a_{ij} = 1$ and $\sum_j a_{ij} \mathbf{v}_j = \mathbf{o}_i$. $Z$ cancels in their ratio:
\begin{equation}
  a'_{ij} = \frac{a_{ij} + u_j}{1 + \sum_k u_k}, \qquad
  \mathbf{o}'_i = \frac{\mathbf{o}_i + \sum_j u_j \mathbf{v}_j}{1 + \sum_j u_j},
\end{equation}
and subtracting $\mathbf{o}_i$ gives \cref{eq:delta-o2}.

\paragraph{Numerics.}
\Cref{eq:delta-o2} cancels when a band holds almost all the attention of a query and a positive score part is removed
($1 + \sum_j u_j \approx 0$). We therefore evaluate $\mathbf{o}'_i$ in fp64 in the equivalent form
\begin{equation}
  \mathbf{o}'_i
  = \frac{\sum_{j \notin \mathcal{B}} a_{ij}\mathbf{v}_j + \sum_{j \in \mathcal{B}} a_{ij} e^{-s^{(l)}_{ij}}\mathbf{v}_j}
         {\sum_{j \notin \mathcal{B}} a_{ij} + \sum_{j \in \mathcal{B}} a_{ij} e^{-s^{(l)}_{ij}}},
\end{equation}
which has only positive summands, and take $\Delta\mathbf{o}_i = \mathbf{o}'_i - \mathbf{o}_i$. We check it against
recomputing the softmax with the score part removed (relative difference below $10^{-4}$, also with a dominant pair). In SmolLM-135M, a few
cells have very large effects (23 of 11\,520 with a ratio above 50): removing a dominant score part moves the
attention to other keys. We therefore report the ratio per cell and shares of cells rather than pooled sums.

\paragraph{Setup.}
The analysis of \cref{sec:interp} uses 256 OpenWebText validation documents of at least 2048 tokens (the first 2048
tokens of each, the same documents for every model) and the distance bands $\{0\}, \{1\}, [2,4), \dots, [1024, 2048)$.
$E^{\text{real}}$ and $E^{\text{null}}$ are summed over the heads of a layer (\cref{eq:e-real}) before taking the
ratio. The null key of a window is the key at the same position in the next window of the batch. The 4K model is
analysed on 2048-token windows so that all models see the same inputs; the initialisation is rebuilt from the seed of
the run, and the intermediate models are weight snapshots saved during training.

\begin{figure}[p]
  \centering
  \includegraphics[width=0.6\textwidth]{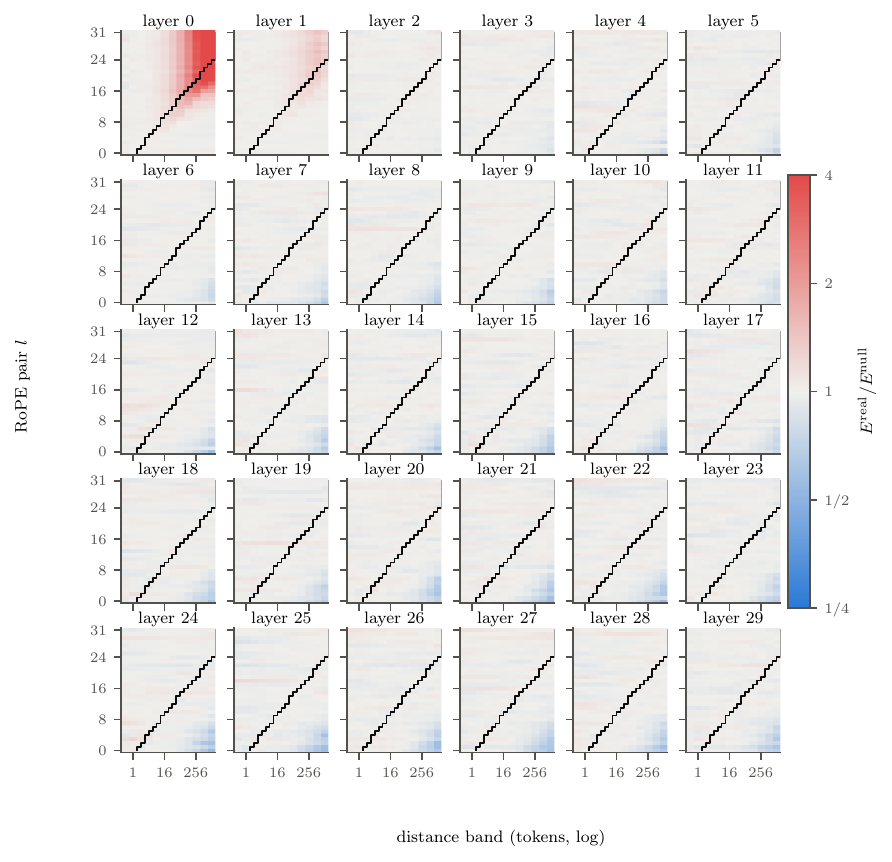}
  \caption{$E^{\text{real}}/E^{\text{null}}$ per layer, pair and distance band, dense 135M model (4K, $b = 10^4$)
    at initialisation. Black line: $\Delta\theta_l = 90^\circ$.}
  \label{fig:interp-init}
\end{figure}

\begin{figure}[p]
  \centering
  \includegraphics[width=0.6\textwidth]{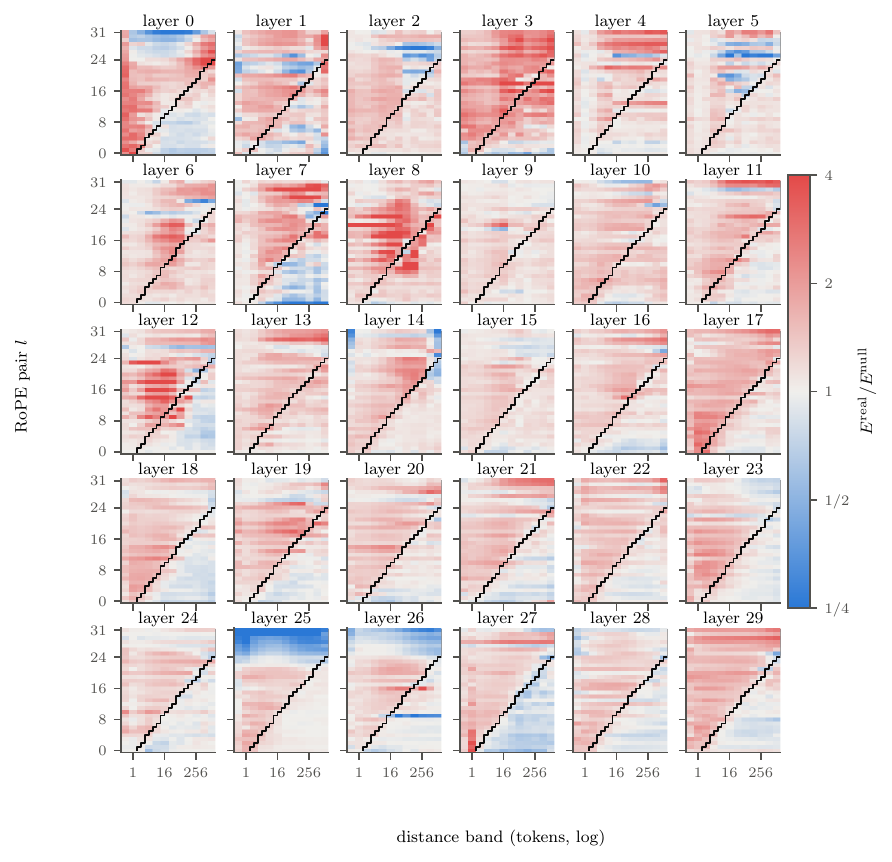}
  \caption{$E^{\text{real}}/E^{\text{null}}$ per layer, pair and distance band, dense 135M model (4K, $b = 10^4$)
    at step 2000. Black line: $\Delta\theta_l = 90^\circ$.}
  \label{fig:interp-2k}
\end{figure}

\begin{figure}[p]
  \centering
  \includegraphics[width=0.6\textwidth]{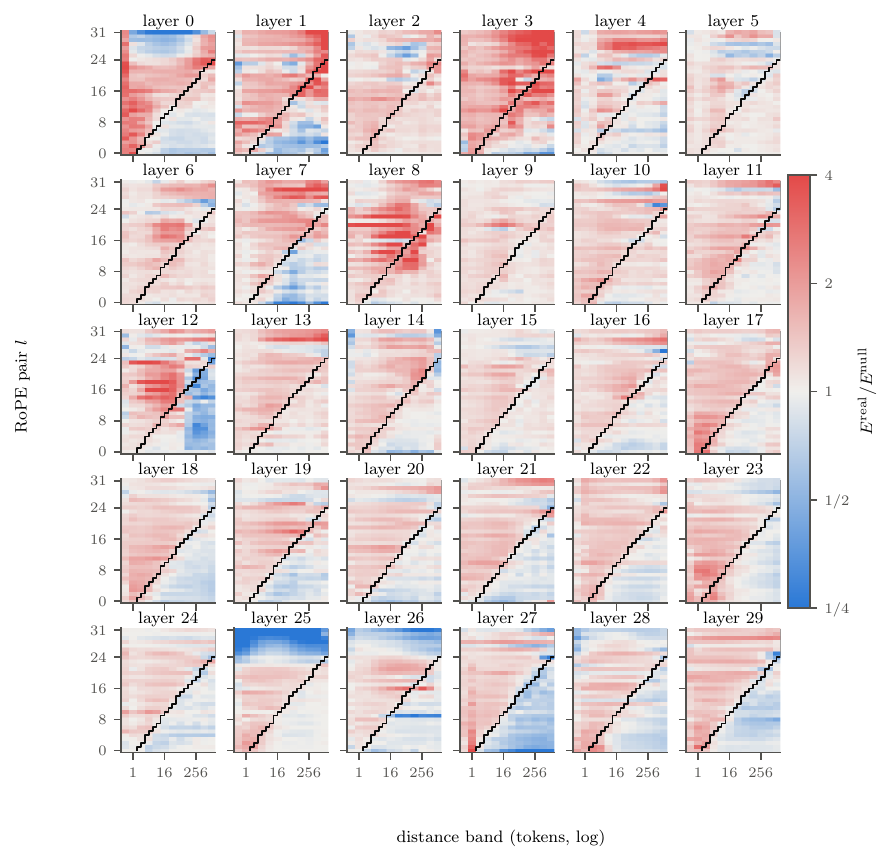}
  \caption{$E^{\text{real}}/E^{\text{null}}$ per layer, pair and distance band, dense 135M model (4K, $b = 10^4$)
    at step 4000. Black line: $\Delta\theta_l = 90^\circ$.}
  \label{fig:interp-4k}
\end{figure}

\begin{figure}[p]
  \centering
  \includegraphics[width=0.6\textwidth]{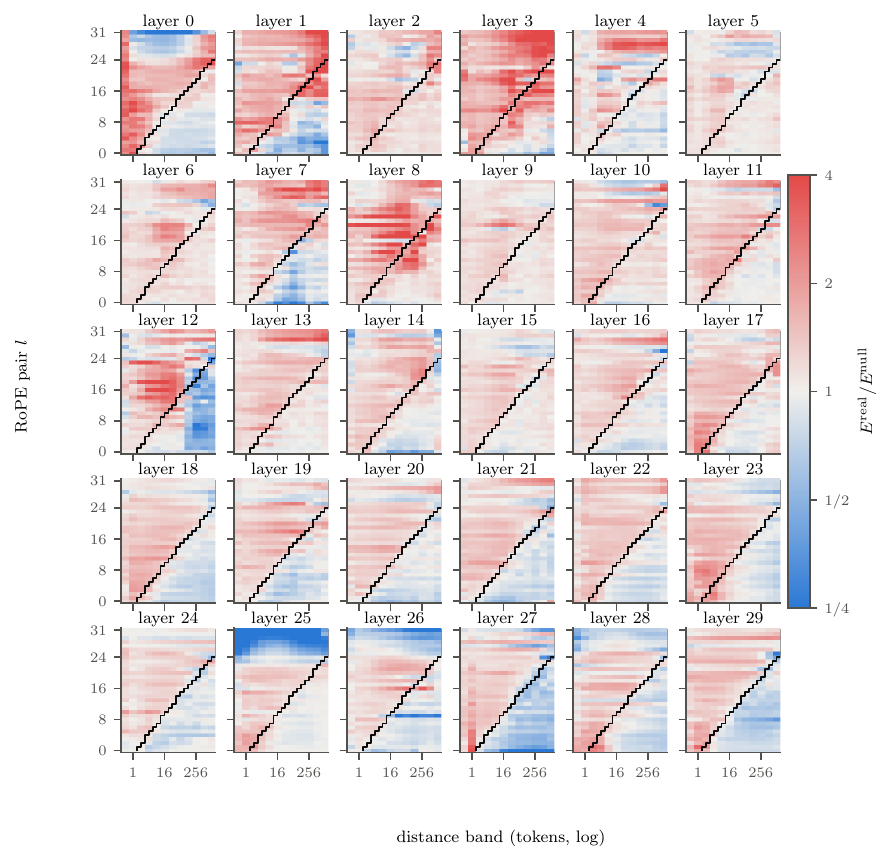}
  \caption{$E^{\text{real}}/E^{\text{null}}$ per layer, pair and distance band, dense 135M model (4K, $b = 10^4$)
    at the end of training (step 5149). Black line: $\Delta\theta_l = 90^\circ$.}
  \label{fig:interp-final}
\end{figure}

\begin{figure}[p]
  \centering
  \includegraphics[width=0.6\textwidth]{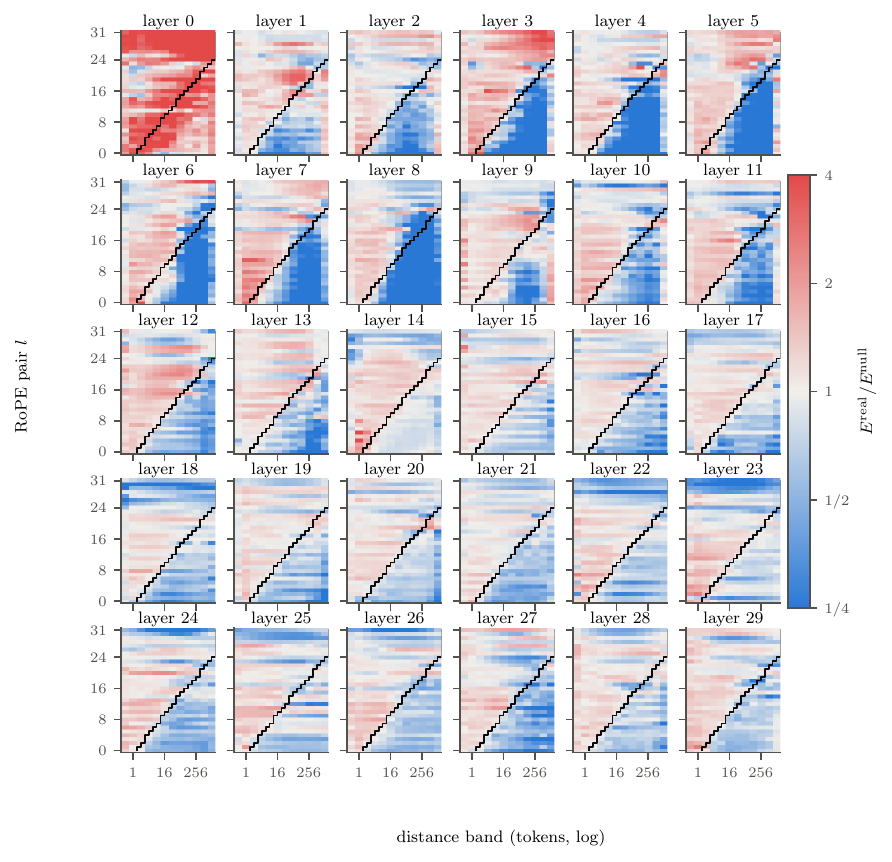}
  \caption{$E^{\text{real}}/E^{\text{null}}$ per layer, pair and distance band, pretrained SmolLM-135M. Black line:
    $\Delta\theta_l = 90^\circ$.}
  \label{fig:interp}
\end{figure}

\paragraph{Observations.}
\Cref{fig:interp-init,fig:interp-2k,fig:interp-4k,fig:interp-final,fig:interp} show all layers. The first layer is an
exception in every model: its score parts are above the reference at every distance, already at initialisation, plausibly
because its keys depend on token identity only and a document repeats its own tokens. At initialisation, the fastest
pairs at the longest distances are slightly below the reference in most layers.

\paragraph{Relative influence.}
The ratio $E^{\text{real}}/E^{\text{null}}$ compares each score part with an uninformative one, but does not show how
large its effect is. We therefore also report its \emph{relative influence}: the change of the head outputs when the
score part is removed, relative to the size of the head outputs themselves,
\begin{equation}
  I_{l,\mathcal{B}} = \frac{\sum_{h} \sum_{i} \big\| W_h \, \Delta\mathbf{o}_{h,i} \big\|^2}
                           {\sum_{h} \sum_{i} \big\| W_h \, \mathbf{o}_{h,i} \big\|^2}
                    = \frac{E^{\text{real}}_{l,\mathcal{B}}}{\sum_{h} \sum_{i} \big\| W_h \, \mathbf{o}_{h,i} \big\|^2},
  \label{eq:relinfl}
\end{equation}
which needs no reference. Both sums run over the same queries and heads as \cref{eq:e-real}; the ratio is taken
between the sums (not averaged over queries), so that queries with a nearly zero output do not dominate.

\begin{figure}[h]
  \centering
  \includegraphics[width=0.5\textwidth]{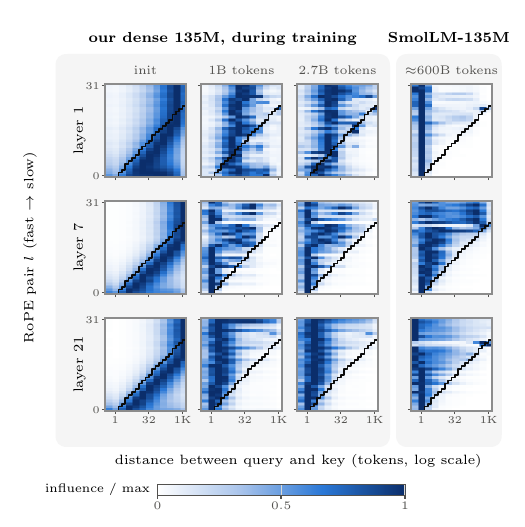}
  \caption{Relative influence $I_{l,\mathcal{B}}$ of each pair (rows of a panel) at each distance band (columns) in
    three layers of the dense 135M model during training and of SmolLM-135M, each pair divided by its maximum over the
    distance bands. Black line: $\Delta\theta_l = 90^\circ$.}
  \label{fig:relinfl-train}
\end{figure}

\begin{figure}[p]
  \centering
  \includegraphics[width=0.6\textwidth]{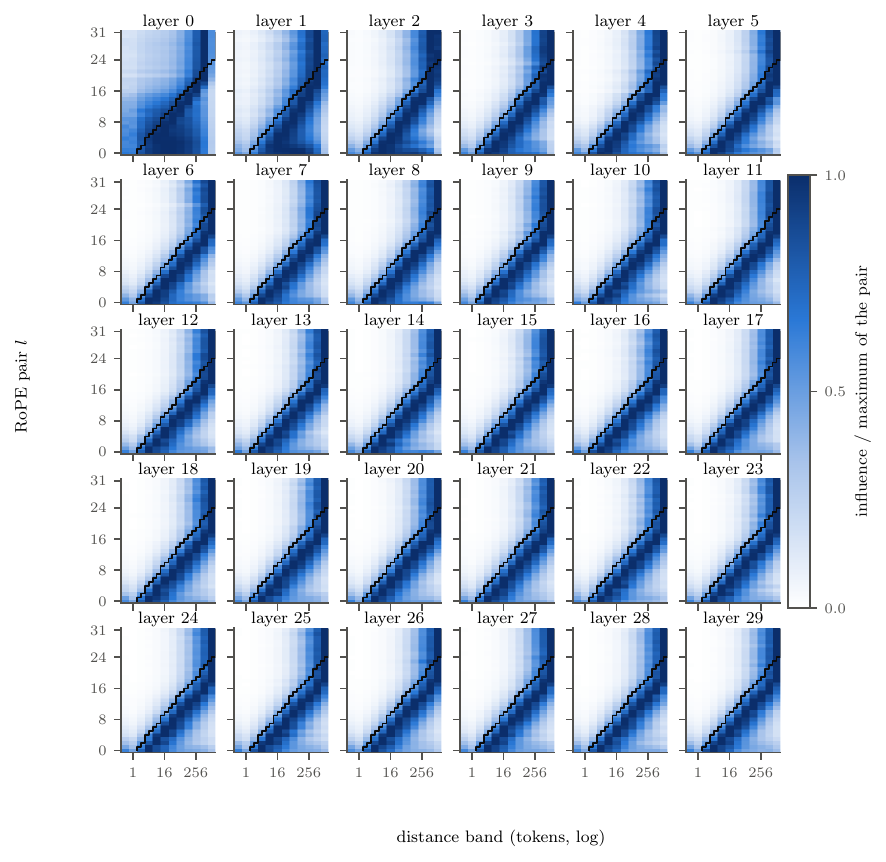}
  \caption{Relative influence $I_{l,\mathcal{B}}$ (\cref{eq:relinfl}) per layer, pair and distance band, dense 135M model (4K, $b = 10^4$) at initialisation;
    each pair divided by its maximum over the distance bands. Black line: $\Delta\theta_l = 90^\circ$.}
  \label{fig:relinfl-init}
\end{figure}

\begin{figure}[p]
  \centering
  \includegraphics[width=0.6\textwidth]{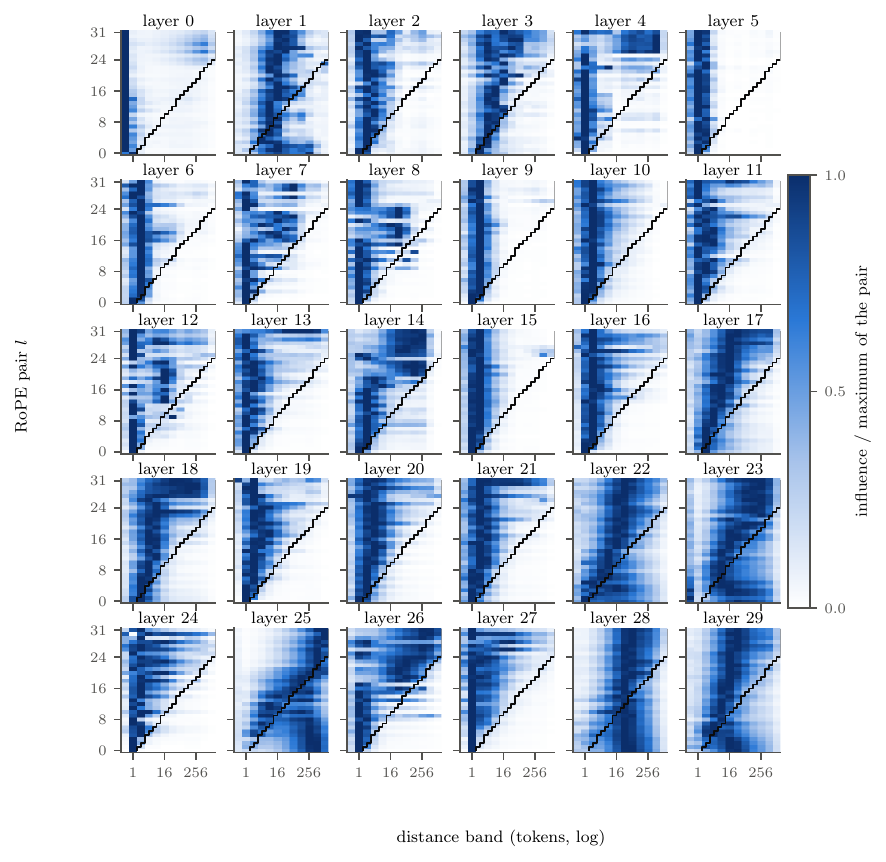}
  \caption{Relative influence $I_{l,\mathcal{B}}$ (\cref{eq:relinfl}) per layer, pair and distance band, dense 135M model (4K, $b = 10^4$) at step 2000;
    each pair divided by its maximum over the distance bands. Black line: $\Delta\theta_l = 90^\circ$.}
  \label{fig:relinfl-2k}
\end{figure}

\begin{figure}[p]
  \centering
  \includegraphics[width=0.6\textwidth]{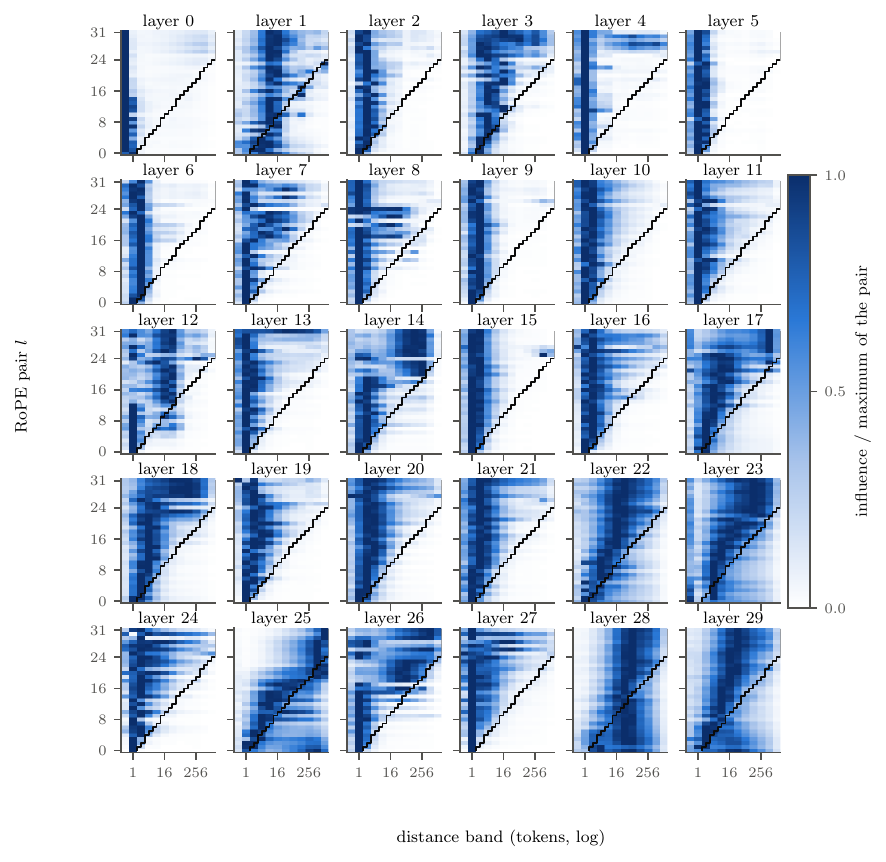}
  \caption{Relative influence $I_{l,\mathcal{B}}$ (\cref{eq:relinfl}) per layer, pair and distance band, dense 135M model (4K, $b = 10^4$) at step 4000;
    each pair divided by its maximum over the distance bands. Black line: $\Delta\theta_l = 90^\circ$.}
  \label{fig:relinfl-4k}
\end{figure}

\begin{figure}[p]
  \centering
  \includegraphics[width=0.6\textwidth]{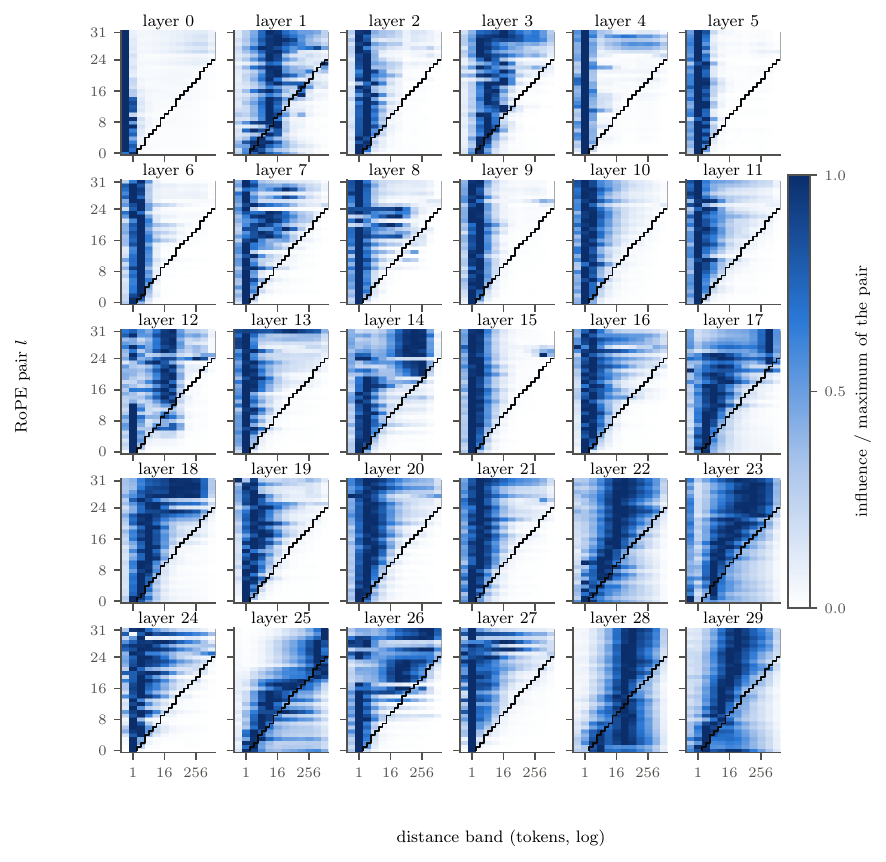}
  \caption{Relative influence $I_{l,\mathcal{B}}$ (\cref{eq:relinfl}) per layer, pair and distance band, dense 135M model (4K, $b = 10^4$) at the end of training (step 5149);
    each pair divided by its maximum over the distance bands. Black line: $\Delta\theta_l = 90^\circ$.}
  \label{fig:relinfl-final}
\end{figure}

\begin{figure}[p]
  \centering
  \includegraphics[width=0.6\textwidth]{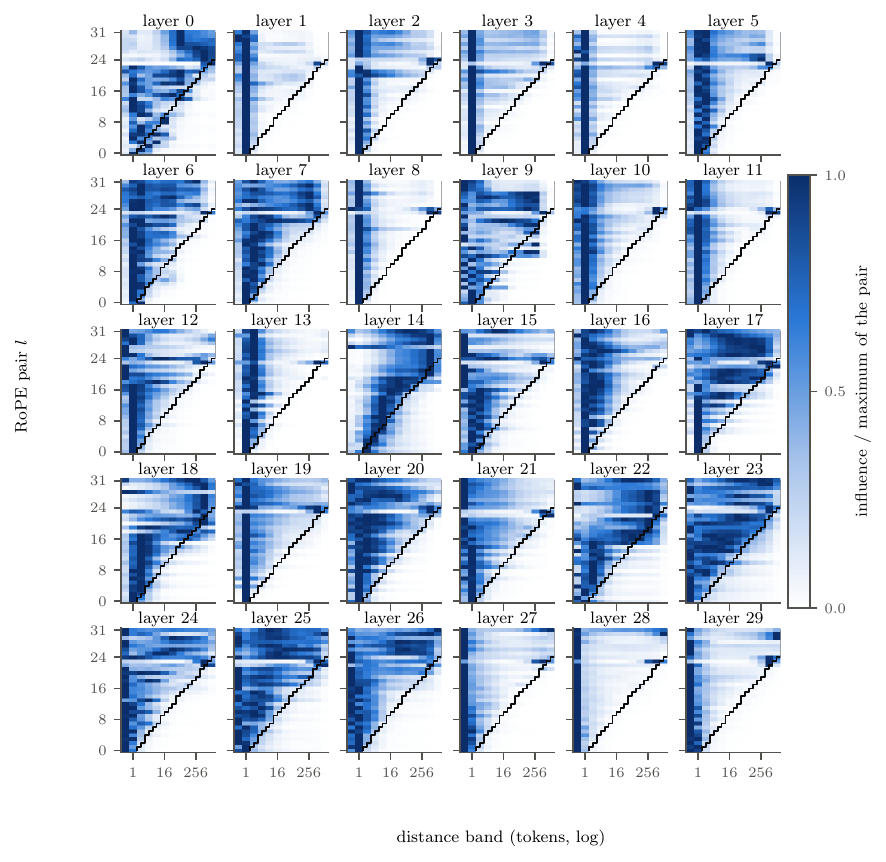}
  \caption{Relative influence $I_{l,\mathcal{B}}$ (\cref{eq:relinfl}) per layer, pair and distance band, pretrained SmolLM-135M;
    each pair divided by its maximum over the distance bands. Black line: $\Delta\theta_l = 90^\circ$.}
  \label{fig:relinfl}
\end{figure}

\Cref{fig:relinfl-train,fig:relinfl-init,fig:relinfl-2k,fig:relinfl-4k,fig:relinfl-final,fig:relinfl} show where the
influence of each pair lies along the distance axis: every pair (row) is divided by its maximum over the distance
bands, so the figures compare the shape of the influence across distances, not its magnitude.

\end{document}